# Autism Disease Detection Using Transfer Learning Techniques: Performance Comparison Between Central Processing Unit vs Graphics Processing Unit Functions for Neural Networks

Mst Shapna Akter
*Dept. of Computer Science*
Kennesaw State University
Kennesaw, USA
Email: makter2@students.kennesaw.edu

Hossain Shahriar
*Dept. of Information Technology*
Kennesaw State University
Kennesaw, USA
Email: hshahria@kennesaw.edu

Alfredo Cuzzocrea
*iDEA Lab*
University of Calabria
Rende, Italy
Email: alfredo.cuzzocrea@unical.it

*Abstract*—Neural network approaches are machine learning methods that are widely used in various domains, such as healthcare and cybersecurity. Neural networks are especially renowned for their ability to deal with image datasets. During the training process with images, various fundamental mathematical operations are performed in the neural network. These operations include several algebraic and mathematical functions, such as derivatives, convolutions, and matrix inversions and transpositions. Such operations demand higher processing power than what is typically required for regular computer usage. Since CPUs are built with serial processing, they are not appropriate for handling large image datasets. On the other hand, GPUs have parallel processing capabilities and can provide higher speed. This paper utilizes advanced neural network techniques, such as VGG16, Resnet50, Densenet, Inceptionv3, Xception, Mobilenet, XGBOOST-VGG16, and our proposed models, to compare CPU and GPU resources. We implemented a system for classifying Autism disease using face images of autistic and non-autistic children to compare performance during testing. We used evaluation matrices such as Accuracy, F1 score, Precision, Recall, and Execution time. It was observed that GPU outperformed CPU in all tests conducted. Moreover, the performance of the neural network models in terms of accuracy increased on GPU compared to CPU.

Keywords: Autism Disease, Neural Network, CPU, GPU, Transfer Learning;

## I. INTRODUCTION

Nowadays, GPU technology efficiently trains and tests deep learning architectures through parallelizable mathematical procedures. GPUs consist of multiple cores that accelerate complex computational operations [1]. The number of processing units (cores) that can operate independently is crucial in determining the level of parallelization possible. There is a substantial difference between GPUs with thousands of cores and CPUs with only four or eight cores [2]. More cores available means an increase in the amount of parallelization possible. When GPUs have a lot of cores, CPU cores operate at a higher frequency. For mathematical procedures in Neural Networks, GPUs are vital [3]. Neural Networks have proven to be very successful in solving several real-life problems. Ensuring high accuracy while working with diseases, such as Autism Spectrum Disorder (ASD), is crucial [4]. ASD brings developmental disability to the brain and is also associated with genetic conditions. Some causes of ASD are known, while others are still unknown [5]. Patients with ASD exhibit different behavior, communication, interaction, and learning styles compared to ordinary people [6]. While some patients can live and work like ordinary people without any support, many patients require assistance from others to live their life. Some patients may have advanced conversation skills, while others may be nonverbal. Usually, ASD starts to develop before the age of three, but some children may show symptoms early in the 12-month period. In the first 12 or 24 months period, they gain knowledge and skills, but later stop learning, making it difficult to communicate with peers and adults, make new friends, and understand complex concepts. People with ASD are also more likely to experience serious issues such as anxiety, depression, and attention-deficit/hyperactivity disorder compared to those without ASD [7]. Early detection of ASD is crucial as it significantly decreases symptoms and has a high chance of improving the quality of life. Medical tests such as blood tests and symptom checkings are the most common ways of detecting ASD. The symptoms are usually checked by parents or teachers, and healthcare providers evaluate the similarity score between possible signs and symptoms provided by the parents.

This study compares the performance of CPU and GPU on different devices analyzed for autism disease using face images. We evaluate well-known deep learning frameworks such as VGG16, Resnet50, Densenet, Inceptionv3, Xception, Mobilenet, XGBOOST-VGG16, and our proposed models using energy-related metrics and offer precise power measurements for each of these frameworks using a collection of carefully chosen networks and layers. We assess various processor architectures, including a Dell Gaming Laptop with an 11th Gen Intel Core i7 operating at 2.30GHz, 16.0 GB of RAM, and 512 GB of SSD storage, as well as a cloud-based service (Google Colab) for testing. We also investigate the impact of various hardware configurations on performance and energy efficiency.

Our contributions can be summarized as follows:

1. We classify Autism disease using advanced Neural Network models such as VGG16, Resnet50, Densenet, Inceptionv3, Xception, Mobilenet, and XGBOOST-VGG16, and show which model performs better on this dataset.

2. We propose our Stacking Neural Network model using the aforementioned six transfer learning models and provide a comparative analysis with traditional models.

3. We conduct a comparison study on the power usage of well-known pre-trained CNN frameworks on various hardware (i.e., Dell Gaming laptop and Google Colab). Our findings provide insights into how the application's

## II. BACKGROUND AND LITERATURE REVIEW

Limited work has been done in analyzing the performance of CPU and GPU on various devices using a Neural Network approach. To gain insight into this, we reviewed various research papers in the field of machine learning and deep learning. In addition, Autism Spectrum Disorder (ASD) is a crucial topic in healthcare, and many researchers have investigated it using various approaches with up-to-date models. We summarize some of the relevant research papers below:

Raj and Masood [8] proposed machine learning techniques for the analysis and detection of autism spectrum disorder. They focused on three age groups: children, adolescents, and adults, using classifiers such as Support Vector Machine (SVM), KNN, Naive Bayes, Logistic Regression, Neural Networks, and Convolutional Neural Networks. Their dataset includes parameters such as nationality, problems at birth, any family member suffering from pervasive development disorders, sex, screening application used by the user before or not, and screening test type. They achieved 99.53%, 98.30%, and 96.88% accuracy for adults, children, and adolescents, respectively. Al-diabat [9] proposed a fuzzy data mining approach for the autism classification of children using the ASDtest app dataset. They compared the performance of fuzzy data mining algorithms (FURIA), JRIP, RIDOR, and PRISM. The accuracy of their proposed models was less than 90 percent. Xie et al. [10] proposed a deep learning model for detecting atypical visual attention in ASD using a two-stream end-to-end network. They used the VGG16 model, including 13 Conv layers, five max-pooling layers, and two fully connected layers, and their proposed model achieved an accuracy of 95%. Liu et al. [11] created a benchmark dataset and proposed multimodal machine learning for ASD detection. The dataset included the spontaneous interaction of adults and children, and the experimental work was carried out using machine learning models such as SVM and RF, achieving 70% accuracy. Duan et al. [12] proposed a system for predicting ASD using human faces and analyzed the visual attention of 300 human faces and their corresponding eye movement images using the CASNET model. De Campos Souza and Guimaraes [13] proposed a fuzzy neural network to predict children with autism and developed a mobile application that uses question and answer-based inputs to generate predictions. Xu et al. [14] proposed child vocalization composition as discriminant information for automatic autism detection, achieving 85% to 90% accuracy using speech datasets. Tao and Shyu et al. [15] proposed a CNN-LSTM-based ASD classification model using observer scan paths, achieving 74.22% accuracy with 300 eye movement datasets for ASD children. Akter et al. [16] proposed a machine learning-based model for early-stage ASD detection for children, adolescents, and adults using SVM, AdaBoost, and GLMboost models. The Adaboost model showed the best performance for children, Glmboost for adolescents, and Adaboost for the adult dataset. Hangun and Eyecioglu [17] analyzed the performance between CPU and GPU resources for image processing operations using Otsu's method and edge detection, finding that GPU resources provide better performance than CPU. Similarly, Buber and Banu [18] compared the performance of CPU and GPU for deep learning techniques using the Recurrent Neural Network (RNN), word embedding, and transfer learning models. They adjusted some hyperparameters to ensure that the models were optimized for each device. The experimental work was carried out on a Tesla k80 GPU and an Intel Xeon Gold 6126 CPU. They found that the GPU was generally faster than the CPU for all three models, with speedup ranging from 1.9x to 3.3x for the RNN and transfer learning models, respectively. The word embedding model showed the highest speedup, with the GPU being 4.2x faster than the CPU. Asano et al. [19] compared the performance of FPGA, GPU, and CPU for image processing tasks. They used three different techniques, including two-dimensional filters, stereo-vision, and k-means clustering, to evaluate the performance of the different devices. They found that while FPGA had a lower operational frequency than CPU and GPU, it could still achieve extremely high performance for image processing tasks. Baykal et al. [20] compared the performance of CPU and GPU on big data using deep learning models, including artificial neural networks (ANN) with a multi-layered structure, deep learning, and convolutional neural networks (CNN). They found that the TensorFlow platform allowed for an extremely effective parallel execution environment on GPUs, reducing execution time by 4 times compared to CPU execution. These studies demonstrate the importance of considering hardware configurations when developing and implementing deep learning models. Understanding the performance of different devices can help researchers and practitioners optimize their models and improve their accuracy, efficiency, and speed.

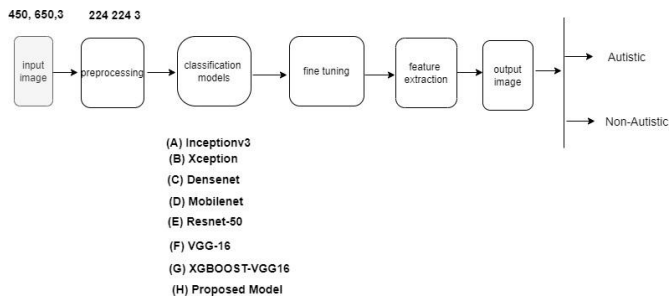

Fig. 1: Proposed Architecture for Autism Disease classification.

## III. MATERIALS AND METHODS

This paper proposes the generalized Neural Network architectures for classifying ASD, shown in Figure 1.

Images are preprocessed using popular techniques by calling the Keras and TensorFlow library and fetched into the classical neural networks such as VGG16, Resnet-50, Mobilenet, Densenet, Inceptionv3, Xception, XGBOOST-VGG16 and our proposed model.

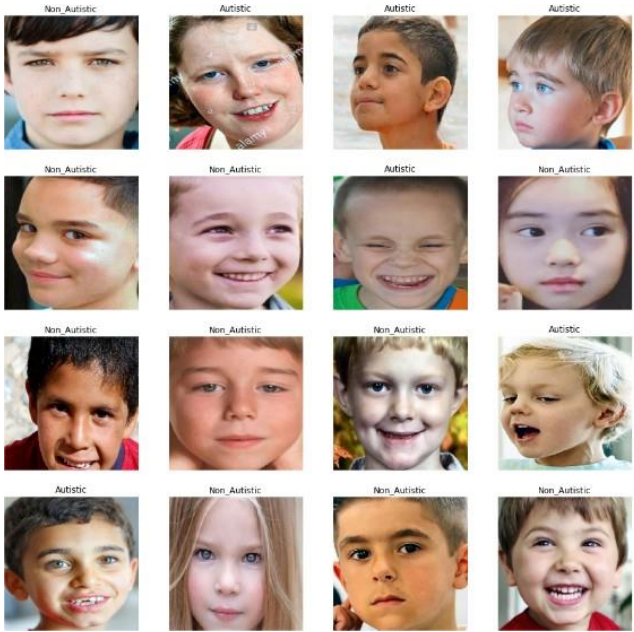

Fig. 2: Sample of Dataset

*A. Dataset*

This paper utilizes a publicly available dataset from a popular Kaggle website, consisting of 2936 facial images accurately divided between ASD and TD children. The contributor collected the dataset through an internet search, stating that they could not obtain ASD images from clinics or verified sources [21]. The original dataset contained more than 3000 images, including incorrect ASD images. Therefore, the data used in this study mainly comprise appropriate data for experimental work [21]. The dataset includes 89% white children and 11% brown/black children, and the purpose of using these images is to efficiently detect ASD disease using facial features.

*B. Preprocessing*

Most image datasets come with improper shapes, which are inappropriate for inputting into Neural Networks. Transfer learning algorithms follow size consistency for each input, which is 224x224 pixels or less. Therefore, we prepared the dataset by transforming it from random sizes to 224x224 pixels, while keeping the channel as 3, and not lowering it below 224x224, as it can negatively impact the model's performance. We retained the noises as they allow the model to learn along with the noises, which helps to accurately recognize real-life ASD.

*C. Classification Models and Fine Tuning*

Convolutional Neural Network (CNN) is a subfield of conventional Neural Network, which comprises three elements: one or more convolutional layers, a pooling layer, and one fully connected layer. The first element is the fundamental building block of the CNN model and calculates the dot product between the kernel and a portion of the input image. The kernel size is relatively smaller than the size of the input image, while the kernel dimension is similar to the input image size. The pooling layer, the second component of the CNN model, aids in reducing computational complexity by excluding some parameters. Lastly, the Softmax function of the fully connected layers provides the probability of input belonging to a particular class.

*1) Inceptionv3:* Szegedy et al. [22] first released Inceptionv3 in 2015. It is a Convolutional Neural Network used to analyze image datasets for tasks such as segmentation, object localization, and image classification [23]. Inceptionv3 is an expanded form of the GoogleNet paradigm. This model reduces the number of parameters that must be trained by concatenating several convolutional filters of varying sizes into a single filter, which eliminates computational complexity. Therefore, classification performance remains good even with fewer parameters. Inceptionv3 has gained popularity among researchers as it can be trained well over large datasets. To fine-tune the model on our dataset, we downloaded the base model and added a GlobalAveragePooling2D layer and a dropout (0.5) model. We also included a dense layer with a "sigmoid" activation, Dropout, and Softmax layers with outputs at the bottom of the architecture. Finally, we fine-tuned the model on 2936 sample images for 30 epochs using the Stochastic gradient descent (SGD) optimizer and a learning rate of 0.0001. The trainable parameters of the Inceptionv3 model are 21,770,401.

*2) Xception:* The Xception model was created by Chollet [24], a Google researcher. Xception is a deep convolutional neural network that uses a linear combination of residual connections and depth-wise separable convolution. The Inception model was used as a foundation for this unique deep neural network design, which substitutes depth-wise separable convolution for the Inception model's inception layers. The architecture is less complex and more effective than existing deep convolutional neural networks due to the idea of developing it with fully depth-wise separable convolution. To further optimize the model, we included the Flatten() layer, BatchNormalization(), a Dense layer with 128 neurons and Relu function, followed by BatchNormalization(), and a final Dense layer with one neuron and sigmoid function, after downloading the base model. Finally, using the Stochastic gradient descent (SGD) optimizer and a learning rate of 0.0001 on 2936 sample images for 30 epochs, we fine-tuned the model. The trainable parameters of the Xception model are 33,853,225.

*3) Densenet:* Gao Huang, Zhuang Liu, and their colleagues introduced the Densely Connected Convolutional Network (Densenet) in 2017 [25]. This deep convolutional neural network connects each layer to every other layer in a feed-forward fashion, taking advantage of dense connections between the layers. Densenet requires fewer parameters for training the model and mitigates the vanishing gradient problem, which makes it ideal for use in computer vision. After downloading the base model, we added the GlobalAveragePooling2D layer and a dropout model (with a rate of 0.5). Additionally, we included a dense layer with "sigmoid" activation, Dropout, and Softmax layers with outputs at the bottom of the architecture to fine-tune the model on the dataset. Finally, using the Stochastic gradient descent (SGD) optimizer and a learning rate of 0.0001 on 2936 sample images for 30 epochs, we fine-tuned the model.

The Densenet model has 6,954,881 trainable parameters.

*4) Mobilenet:* The MobileNet architecture was first introduced by Andrew G. Howard et al. [26]. It is a shallow deep neural network model that is good for reducing computation time and cost. MobileNet substitutes depthwise separable convolutions for conventional convolutions to minimize the model size and processing. Using a factorization technique called depthwise separable convolution, a normal convolution is split into two convolutions: a depthwise convolution and a pointwise convolution. A 1x1-dimensional convolution is a pointwise convolution. While pointwise convolution attempts to integrate, depthwise convolution aims to filter. The process of depthwise separable convolution is the result of adding depthwise and pointwise convolution. Except for the top layer, MobileNet's architecture is composed of separable convolutions. After downloading the base model, we included the GlobalAveragePooling2D layer and a dropout (0.5) model. To further fine-tune the model on the dataset, we included a dense layer with "sigmoid" activation, Dropout, and Softmax layers with outputs at the bottom of the architecture. Finally, we fine-tuned the model using the Stochastic gradient descent (SGD) optimizer and a learning rate of 0.0001 on 2936 sample images for 30 epochs. The trainable parameters of the MobileNet model are 3,208,001.

*5) Resnet-50:* Residual Neural Networks (ResNet) are a type of Artificial Neural Network (ANN) that stack leftover blocks to create a network. The ResNet-34, ResNet-50, and ResNet-101 are the most well-known ResNet networks, with variations distinguished by the number of layers. ResNet-50, for instance, employs 50 layers of neural networks. Using many layers to address complex issues is beneficial because each layer can be trained for a variety of activities. However, the deeper the network, the more likely it is to encounter a degradation issue, which is problematic. The initialization of the network, an optimization algorithm, or a problem with vanishing or exploding gradients typically causes this problem. The ResNet model seeks to prevent these problems by using skip connections, which are at the center of the residual blocks. Skip connections are the strength of the ResNet model, and two methods are used. First, they create a different route around the gradient to mitigate the vanishing gradient problem. Second, they give the model the opportunity to learn an identity function that ensures that the upper levels function nearly identically to the lower layers. The model can successfully classify the limited dataset since it has been trained on more than a million images. After downloading the base model, we included the Flatten() layer, BatchNormalization(), a Dense layer with 128 neurons and Relu function, again BatchNormalization(), and a final Dense layer with one neuron and sigmoid function. Lastly, using the Stochastic gradient descent (SGD) optimizer and a learning rate of 0.0001 on 2936 sample images for 30 epochs, we fine-tuned the model. The trainable parameters of the ResNet-50 model are 23,796,993.

*6) VGG-16:* VGG-16 is the most advanced deep neural network for interpreting visual input. It is considered one of the best architectures for vision models to date, and it won the 2014 ILSVR (Imagenet) competition. The network has a whopping 138 million parameters, and its name comes from the fact that it has 16 weighted layers. VGG16's distinctive feature is that it retains a convolution layer of 3x3 filters with a stride 1 and the same padding and a max pool layer of 2x2 filters with a stride 2 throughout the entire architecture. The model's final two layers consist of two FC (fully connected) layers, followed by a softmax as the output. After downloading the base model, we included the GlobalMaxPooling2D(), a Dense layer with 512 neurons followed by a Relu function, a Dropout(0.5) layer, and a final Dense layer with one neuron and sigmoid function. Lastly, using the Stochastic gradient descent (SGD) optimizer and a learning rate of 0.0001 on 2936 sample images for 30 epochs, we fine-tuned the model. The trainable parameters of the VGG16 model are 14,977,857.

*7) Proposed model:* We utilized the stacking ensemble learning approach to develop our proposed system, which involved logistic regression as the meta-learner and heterogeneous weak learners as the base models. Inceptionv3, Xception, Densenet, Mobilenet, Resnet-50, and VGG-16 were used as the heterogeneous base models, which took in three-dimensional images as inputs. The data separation process differed from that of traditional models, with 60% of the data being used for training, 10% for validation, and 30% for testing. This modification was necessary to avoid overfitting during the meta-learner training phase. The predicted dataset from Level 0 already contained a probability of expected values, enabling the meta-learner to provide accurate probabilities from Level 0. To prevent overfitting, the final model (meta-learner) was trained using both the validation dataset and the outputs. The level-1 prediction was the ultimate result. The stacking ensemble model produced a final model (meta-learner) by using the predicted results from several other models. The disadvantage of the typical stacking ensemble method is that it employs the same various models for the base model, which leads to similar estimates. If the basic model performs poorly on the dataset, there is a high possibility that the final output will be inferior. However, the single neural network model demonstrates bias and volatility toward the dataset. As a result, different models were chosen when constructing the base model.

The architecture is divided into two levels: Level 0 and Level 1, as shown in Figure 9. Inceptionv3, Xception, Densenet, Mobilenet, Resnet-50, and VGG-16 models are used to construct Level 0. The six submodels each make six predictions concurrently after learning the data pattern. Each model employed in Level 0 contributes equally to the overall model. Logistic regression is used to build Level 1, also known as the meta-learner. The meta-learner at Level 1 is fed the Level 0 projected outputs as input. Based on the anticipated Level 0 outputs, the meta-learner estimates the weighted outputs best. A model that can quickly learn a pattern or adjust to different datasets with a small amount of training data is referred to as a "meta learner." It learns patterns of the outputs derived from the six models. As a result, the model can learn completely new data quite effectively and produce acceptable output. The parameter of this model is a combination of six transfer learnings' learnable parameters.

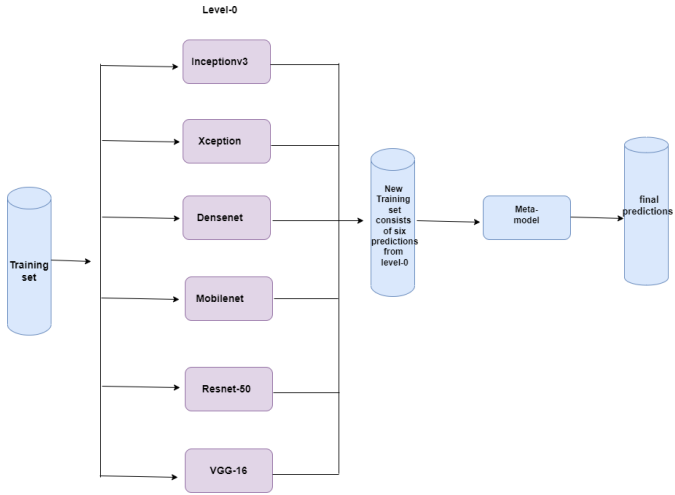

Fig. 3: Proposed Architecture for Autism Disease classification.

*8) XGBOOST-VGG16:* At the beginning of our process, we effectively extracted the desired features using the traditional VGG16 model. We then used a popular classifier called XGBoost to classify the features. Extreme Gradient Boosting (XGBoost) is a distributed, scalable gradient-boosted decision tree (GBDT) machine learning framework. It offers parallel tree boosting for regression, classification, and ranking issues. In supervised machine learning, a model is trained using algorithms to discover patterns in a dataset of features and labels, and the model is then used to predict the labels on the features of a new dataset. Gradient boosting is a supervised learning approach that combines an ensemble of estimates from a variety of weaker models to anticipate a target variable appropriately. The XGBoost algorithm outperforms machine learning challenges because of its efficient handling of a wide range of data types, relationships, distributions, and adjustable hyperparameters. Regression, classification (binary and multiclass), and ranking issues can all be addressed using XGBoostIt is important to evaluate the performance of a model to understand how well the predicted results match the actual ones [27]. Evaluation metrics are used to measure a model's performance, but the choice of metrics depends on the type of model [28]. There are two types of models: classification and regression. Regression is used when predicting a numerical value, while classification is used when predicting a discrete value. The error measure is used to evaluate regression models [29], while classification models are evaluated using the accuracy metric. In our study, we aim to classify ASD patterns of face shape [30], and therefore, we utilized accuracy, F1 score, precision, and recall as our evaluation metrics.

*D. Evaluation Metrics*

It is important to evaluate the performance of a model to understand how well the predicted results match the actual ones [27]. Evaluation metrics are used to measure a model's performance, but the choice of metrics depends on the type of model [28]. There are two types of models: classification and regression. Regression is used when predicting a numerical value [31], while classification is used when predicting a discrete value [32?, 33]. The error measure is used to evaluate regression models [29], while classification models are evaluated using the accuracy metric. In our study, we aim to classify ASD patterns of face shape [30], and therefore, we utilized accuracy, F1 score, precision, and recall as our evaluation metrics.

**Precision**: Precision measures how accurately the model predicts a positive value when it is actually positive. It is used when False Positives are prevalent. If the model's classification accuracy is low, many non-ASD images will be mistakenly identified as having ASD. Conversely, if the accuracy is high, the model will learn from false alarms and ignore the False Positive results. The following formula is used to calculate precision:

$$\text{Precision} = \frac{TP}{TP + FP} \quad (1)$$

Here, TP refers to True Positive values, and FP refers to False Positive values.

**Recall**: Recall is the inverse of Precision. It measures how well the model correctly identifies a positive value when it is actually positive. It is used when the False Negatives (FN) are high. If the model provides limited recall, many ASD images will be classified as non-ASD in the ASD classification problem. Conversely, if it provides high recall, the model will learn from false alarms and disregard the false negative results. The following formula is used to calculate recall:

$$\text{Recall} = \frac{TP}{TP + FN} \quad (2)$$

**F1 score**: The F1 score combines Precision and Recall to determine the model's overall accuracy. The F1 score ranges from 0 to 1. The F1 score returns 1 if the predicted value agrees with the expected value and 0 if none of the values agree with the expected value. The following formula is used to calculate the F1 score:

$$\text{F1 score} = \frac{2 \cdot \text{precision} \cdot \text{recall}}{\text{precision} + \text{recall}} \quad (3)$$

**Accuracy**: Accuracy measures how well the predicted output matches the actual value.

$$\text{Accuracy} = \frac{TP + TN}{TP + TN + FP + FN} \quad (4)$$

Here, TN refers to True Negative, and FN refers to False Negative.

*E. Experimental settings*

This paper uses two devices: physical device and online device. We referred physical device as device 1, and online device as device 2.

*1) Device 1:* **Device Specification details:** Device name: Dell - G15 15.6" FHD 120Hz Gaming Laptop Processor: 11th Gen Intel(R) Core(TM) i7-11800H @ 2.30GHz 2.30 GHz Installed RAM: 16.0 GB (15.7 GB usable) System type: 64-bit operating system, x64-based processor Memory: 512GB Solid State Drive GPU: NVIDIA GeForce RTX 3050

**GPU Configuration:** NVIDIA released the GeForce RTX 3050 8 GB on January 4, 2022, as a performance-segment graphics card [34]. The card supports DirectX 12 Ultimate and is based on the GA106 graphics processor in its GA106-150-KA-A1 variant. It is manufactured using an 8 nm process. This guarantees that GeForce RTX 3050 8 GB will be able to run all current games. Future video games will include hardware raytracing, variable-rate shading, and other features due to the DirectX 12 Ultimate capabilities. The GA106 graphics processor has a die area of 276 mm² and 12,000 million transistors, making it a chip of ordinary size [35]. NVIDIA has disabled some shading units on the GeForce RTX 3050 8 GB to reach the product's target shader count, in contrast to the fully unlocked GeForce RTX 3060 8 GB, which uses the same GPU but has all 3584 shaders activated. It has 32 ROPs, 80 texture mapping units, and 2560 shading units. Eighty tensor cores are also present, which help machine learning applications run faster. Twenty acceleration cores for raytracing are included on the GPU. The GeForce RTX 3050 8 GB and 8 GB of GDDR6 RAM from NVIDIA are connected via a 128-bit memory interface. Memory is running at 1750 MHz, and the GPU is functioning at 1552 MHz, with a boost to 1777 MHz (14 Gbps effective). The NVIDIA GeForce RTX 3050 8 GB is a dual-slot card that uses one 8-pin power connector with a maximum power draw of 130 W. 1x HDMI 2.1 and 3x DisplayPort 1.4a are available as display outputs. A PCI-Express 4.0 x8 interface links the GeForce RTX 3050 8 GB to the rest of the hardware. The card has a dual-slot cooling system and measures 242 mm long by 112 mm wide [35].

*2) Device2:* Google Colaboratory (CoLab) is an open-source platform provided by Google Research. Colab is particularly well-suited to machine learning, data analysis, and education. It enables anyone to create and execute arbitrary Python code through the browser [36]. Moreover, Colab is a hosted Jupyter notebook service that requires no installation and offers free access to computer resources such as GPUs [37].

**Device Specification details:**

Device name: Google Colab CPU: Intel (R) Xeon (R) CPU @ 2.20GHz Memory: 12 GB Storage: 86GB GPU: Tesla T4

**GPU Configuration:**

NVIDIA introduced the Tesla T4 professional graphics card on September 13, 2018. The card supports DirectX 12 Ultimate and is based on the TU104 graphics processor in its TU104-895-A1 model. It was manufactured using a 12 nm processor with 13,600 million transistors and a die area of 545 mm². The TU104 graphics engine is a large chip. NVIDIA has disabled some shading units on the Tesla T4 to reach the product's target shader count, in contrast to the fully unlocked GeForce RTX 2080 SUPER, which utilizes the same GPU but has all 3072 shaders enabled. The Tesla T4 has 64 ROPs, 160 texture mapping units, and 2560 shading units. It also includes 320 tensor cores, which accelerate machine learning applications, and forty acceleration cores for raytracing. The Tesla T4 is coupled with 16 GB of GDDR6 memory from NVIDIA using a 256-bit memory interface. The memory runs at 1250 MHz, while the GPU operates at 585 MHz and can be clocked up to 1590 MHz (10 Gbps effective) [38]. The NVIDIA Tesla T4 is a single-slot card that does not require a separate power connector, and its maximum power usage is 70 W. As it is not intended to have monitors connected to it, this device lacks display connectivity. The Tesla T4 is linked to the rest of the hardware through a PCI-Express 3.0 x16 interface. The card has a single-slot cooling system and measures 168 mm in length [39].

## IV. RESULTS AND DISCUSSIONS

We have shown results derived from Neural Network models, such as Inceptionv3, Xception, Densenet, Mobilenet, Resnet-50, VGG-16, XGBoost-VGG-16, and our proposed model with parameters 21,770,401, 6,954,881, 3,208,001, 23,796,993, and 14,977,857, respectively. Since the trainable parameters of each model are different, the accuracy and execution time differ from model to model. We used evaluation metrics such as Accuracy, Precision, Recall, and F1 Score. All the experiments have been carried out on two devices with and without GPU. The two devices are: Dell - G15 15.6" FHD 120Hz Gaming Laptop and Google Colab. We referred to the Dell - G15 15.6" FHD 120Hz Gaming Laptop as Device 1 and Google Colab as Device 2. Moreover, we denoted Device 1 with GPU as $D_1$, Device 1 without GPU as $D_1'$, Device 2 with GPU as $D_2$, and Device 2 without GPU as $D_2'$. Table 1, Table 2, Table 3, and Table 4 show classification results derived from different Neural Network models for the two devices with and without GPU. From the experiments, we found that while running the Neural Networks on Device 1, the execution time is comparatively lesser than on Device 2, both in the case of GPU support and without GPU support. Moreover, GPU accelerates the execution time significantly for both devices. For Device 1 without GPU, VGG16, Resnet50, Densenet, Inceptionv3, Xception, Mobilenet, XGBOOST-VGG16, and our proposed model took 5h 27min, 3h 2min, 4h 42min, 2h 9min, 3h 20min, 24min 29s, 31min 23s, and 6h 55min, respectively. While on Device 1 with GPU, VGG16, Resnet50, Densenet, Inceptionv3, Xception, Mobilenet, XGBOOST-VGG16, and our proposed model took 33min 42s, 25min 9s, 23min 5s, 18min 30s, 25min 33s, 5min 58s, 8min 49s, and 45min 22s, respectively. Therefore, Device 1 with GPU support takes 4h 54min, 2h 6min, 4h 19min, 1h 51min, 2h 55min, 19min, and 6h 10min less time than VGG16, Resnet50, Densenet, Inceptionv3, Xception, Mobilenet, XGBOOST-VGG16, and our proposed model. Moreover, the accuracy also increases when training the Neural Networks with GPU. For Device 1 without GPU, VGG16, Resnet50, Densenet, Inceptionv3, Xception, Mobilenet, XGBOOST-VGG16, and our proposed model provide an accuracy of 0.87, 0.90, 0.89, 0.92, 0.90, 0.89, 0.95, and 0.93, respectively.

TABLE I: Autism Disease Classification Results Training Different Neural Network Models Without GPU support for Device1

| Models | Accuracy | precision | Recall | F1 Score | Execution Time |
|---|---|---|---|---|---|
| VGG16 | 0.87 | 0.90 | 0.91 | 0.92 | 5h 27min |
| Resnet50 | 0.90 | 0.98 | 1.00 | 0.92 | 3h 2min |
| Densenet | 0.89 | 0.91 | 0.92 | 0.90 | 4h 42min |
| Inceptionv3 | 0.92 | 0.99 | 0.94 | 0.91 | 2h 9min |
| Xception | 0.90 | 0.90 | 1.00 | 0.92 | 3h 20min |
| Mobilenet | 0.89 | 0.90 | 0.91 | 0.87 | 24min 29s |
| XGBOOST-VGG16 | 0.95 | 1.00 | 1.00 | 0.99 | 31min 23s |
| Proposed Model | 0.93 | 0.99 | 1.00 | 0.97 | 6h 55min |

TABLE II: Autism Disease Classification Results Training Different Neural Network Models With GPU support for Device1

| Models | Accuracy | precision | Recall | F1 Score | Execution time |
|---|---|---|---|---|---|
| VGG16 | 0.89 | 0.90 | 0.91 | 0.93 | 33 min 42s |
| Resnet50 | 0.93 | 0.94 | 0.93 | 0.96 | 25min 9s |
| Densenet | 0.92 | 0.91 | 0.95 | 0.90 | 23min 5s |
| Inceptionv3 | 0.94 | 0.92 | 0.95 | 0.92 | 18min 30s |
| Xception | 0.90 | 0.90 | 0.99 | 0.95 | 25min 33s |
| Mobilenet | 0.91 | 0.90 | 0.94 | 0.93 | 5min 58s |
| XGBOOST-VGG16 | 0.94 | 0.98 | 0.97 | 0.99 | 8min 49s |
| Proposed Model | 0.97 | 0.99 | 1.00 | 0.99 | 45min 22s |

TABLE III: Autism Disease Classification Results Training Different Neural Network Models Without GPU support for Device2

| Models | Accuracy | precision | Recall | F1 Score | Execution Time |
|---|---|---|---|---|---|
| VGG16 | 0.82 | 0.87 | 0.80 | 0.80 | 8h 7min |
| Resnet50 | 0.86 | 0.77 | 0.93 | 0.92 | 8h 22min |
| Densenet | 0.79 | 0.78 | 0.82 | 0.90 | 7h 19min |
| Inceptionv3 | 0.90 | 0.89 | 0.98 | 0.98 | 7h 27min |
| Xception | 0.88 | 0.80 | 0.92 | 0.93 | 8h 29min |
| Mobilenet | 0.82 | 0.88 | 0.90 | 0.91 | 2h 15min |
| XGBOOST-VGG16 | 0.93 | 0.91 | 0.92 | 0.93 | 52min 1s |
| Proposed Model | 0.92 | 0.94 | 0.97 | 0.97 | 9h 18min |

TABLE IV: Autism Disease Classification Results Training Different Neural Network Models With GPU support for Device2

| Models | Accuracy | precision | Recall | F1 Score | Execution time |
|---|---|---|---|---|---|
| VGG16 | 0.87 | 0.90 | 0.89 | 0.85 | 41 min 38s |
| Resnet50 | 0.89 | 0.92 | 0.90 | 0.92 | 35min 19s |
| Densenet | 0.88 | 0.85 | 0.86 | 0.93 | 31min 41s |
| Inceptionv3 | 0.83 | 0.87 | 0.88 | 0.89 | 19min 1s |
| Xceptionv3 | 0.87 | 0.88 | 0.89 | 0.92 | 32min 9s |
| Mobilenet | 0.83 | 0.86 | 0.88 | 0.88 | 12min 2s |
| XGBOOST-VGG16 | 0.93 | 0.91 | 0.95 | 0.97 | 19min 1s |
| Proposed Model | 0.92 | 0.97 | 0.99 | 0.95 | 1hr 3min |

## V. CONCLUSION

In this modern era, the concept of Neural Networks has become highly demanded for dealing with image datasets. It is also used for applications requiring text datasets, numerical values, etc. Neural Networks are known for their accurate prediction and classification, making them crucial for resolving many real-life issues. However, training Neural Networks on large datasets can be time-consuming. To address this issue, accelerating technologies such as parallel computation on multi-core platforms like GPUs are employed. In this study, we utilized CUDA-supported CPU and GPU functions to measure the time spent on widely used neural network techniques. We also focused on recognizing the autistic disease and demonstrated efficient results using a dataset consisting of images of autistic children's faces. Furthermore, we investigated the impact of different CPU and GPU hardware configurations on the functionality and energy usage of CNNs. Our findings show that GPUs outperform CPUs in terms of accuracy and required time in all cases. These findings can be applied to the development of energy-efficient neural network designs and deep CNN frameworks. The measurements reveal improved performance due to the parallelized nature of GPU functions.